\documentclass[twocolumn,journal]{IEEEtran}

\usepackage[T1]{fontenc}
\usepackage[latin9]{inputenc}
\usepackage{color}
\usepackage{url}
\usepackage{amsmath}
\usepackage{amssymb}
\usepackage{graphicx}
\usepackage[unicode=true,
 bookmarks=true,bookmarksnumbered=true,bookmarksopen=true,bookmarksopenlevel=1,
 breaklinks=false,pdfborder={0 0 0},pdfborderstyle={},backref=false,colorlinks=false]
 {hyperref}
\hypersetup{pdftitle={Your Title},
 pdfauthor={Your Name},
 pdfpagelayout=OneColumn, pdfnewwindow=true, pdfstartview=XYZ, plainpages=false}

\makeatletter
\let\oldforeign@language\foreign@language
\DeclareRobustCommand{\foreign@language}[1]{%
  \lowercase{\oldforeign@language{#1}}}

\usepackage[caption=false,font=footnotesize]{subfig}
\usepackage{xcolor}
\usepackage{graphicx}
\usepackage{cite}
\usepackage{balance}

\definecolor{lightgray3}{rgb}{0.9,0.9,0.9}

\makeatother

\usepackage{listings}
\begin{document}
\title{A Multi-Arm Robotic Platform for Scientific Exploration}
\author{Murilo~Marques~Marinho,~\IEEEmembership{Member,~IEEE}, Juan~José~Quiroz-Omaña,~and
Kanako~Harada,~\IEEEmembership{Member,~IEEE} \thanks{This project was funded by JST Moonshot R\&D (JPMJMS2033-09).}\thanks{M. M. Marinho is with the University of Tokyo, Department of Mechanical
Engineering, Tokyo, Japan. e-mail: \protect\href{mailto:murilomarinho@ieee.org}{murilomarinho@ieee.org}.}\thanks{J. J. Q. Omaña and K. Harada are with the University of Tokyo, Graduate
Schools of Medicine \& Engineering, Tokyo, Japan. e-mails: \protect\href{mailto:\%7Bjuanjqo,kanakoharada\%7D@g.ecc.u-tokyo.ac.jp}{\{juanjqo,kanakoharada\}@g.ecc.u-tokyo.ac.jp}.}}
\markboth{ACCEPTED FOR PUBLICATION IN THE IEEE ROBOTICS AND AUTOMATION MAGAZINE}{Marinho, Omaña, and Harada: A Multi-Arm AI Platform for Scientific
Exploration}
\maketitle
\begin{abstract}
There are a large number of robotic platforms with two or more arms
targeting surgical applications. Despite that, very few groups have
employed such platforms for scientific exploration. Possible applications
of a multi-arm platform in scientific exploration involve the study
of the mechanisms of intractable diseases by using organoids, i.e.,
miniature human organs) The study of organoids requires the preparation
of a cranial window, which is done by carefully removing an 8 mm patch
of the skull of a mouse. In this work, we present the first prototype
of our AI-robot science platform for scientific experimentation, its
digital twins, and perform validation experiments under teleoperation.
The experiments showcase the dexterity of the platform by performing
peg transfer, gauze cutting, mock experiments using eggs, and the
world's first four-hand teleoperated drilling for a cranial window.
\textcolor{blue}{The digital twins and related control software are
freely available for noncommercial use at \url{https://AISciencePlatform.github.io}.}
\end{abstract}

\global\long\def\dq#1{\underline{\boldsymbol{#1}}}%

\global\long\def\quat#1{\boldsymbol{#1}}%

\global\long\def\mymatrix#1{\boldsymbol{#1}}%

\global\long\def\myvec#1{\boldsymbol{#1}}%

\global\long\def\mapvec#1{\boldsymbol{#1}}%

\global\long\def\dualvector#1{\underline{\boldsymbol{#1}}}%

\global\long\def\dual{\varepsilon}%

\global\long\def\dotproduct#1{\langle#1\rangle}%

\global\long\def\norm#1{\left\Vert #1\right\Vert }%

\global\long\def\mydual#1{\underline{#1}}%

\global\long\def\hamilton#1#2{\overset{#1}{\operatorname{\mymatrix H}}\left(#2\right)}%

\global\long\def\hamiquat#1#2{\overset{#1}{\operatorname{\mymatrix H}}_{4}\left(#2\right)}%

\global\long\def\hami#1{\overset{#1}{\operatorname{\mymatrix H}}}%

\global\long\def\tplus{\dq{{\cal T}}}%

\global\long\def\getp#1{\operatorname{\mathcal{P}}\left(#1\right)}%

\global\long\def\getd#1{\operatorname{\mathcal{D}}\left(#1\right)}%

\global\long\def\swap#1{\text{swap}\{#1\}}%

\global\long\def\imi{\hat{\imath}}%

\global\long\def\imj{\hat{\jmath}}%

\global\long\def\imk{\hat{k}}%

\global\long\def\real#1{\operatorname{\mathrm{Re}}\left(#1\right)}%

\global\long\def\imag#1{\operatorname{\mathrm{Im}}\left(#1\right)}%

\global\long\def\imvec{\boldsymbol{\imath}}%

\global\long\def\vector{\operatorname{vec}}%

\global\long\def\mathpzc#1{\fontmathpzc{#1}}%

\global\long\def\cost#1#2{\underset{\text{#2}}{\operatorname{\text{cost}}}\left(\ensuremath{#1}\right)}%

\global\long\def\diag#1{\operatorname{diag}\left(#1\right)}%

\global\long\def\frame#1{\mathcal{F}_{#1}}%

\global\long\def\ad#1#2{\text{Ad}\left(#1\right)#2}%

\global\long\def\adsharp#1#2{\text{Ad}_{\sharp}\left(#1\right)#2}%

\global\long\def\spin{\text{Spin}(3)}%

\global\long\def\spinr{\text{Spin}(3){\ltimes}\mathbb{R}^{3}}%

\global\long\def\argminimtwo#1#2#3#4#5{ \begin{aligned}#1\:  &  \underset{#2}{\arg\!\min}  &   &  #3 \\
  &  \text{subject to}  &   &  #4\\
  &   &   &  #5 
\end{aligned}
 }%

\section{Introduction}

\IEEEPARstart{T}{he collaboration} among humans and robots is a reality
in many fields of industry, science, and medicine. The robotics community
has made much progress in the direction of making that collaboration
safe and meaningful. We often rely on robots for tasks that might
be over the limit of human accuracy or endurance, and humans for decision
making and higher-level recognition.

This development has been further empowered by the recent re-rise
of data-driven \emph{deep learning }\cite{LeCun_2015} (usually referred
to as \emph{artificial intelligence}\textendash AI), which has been
undoubtedly influential in robotics research since at least 2015 \cite{S_nderhauf_2018}.
The success of these technologies has sparked the interest of society
in the impact of future AI-embedded robots (henceforth \emph{AI-robots}),
further rekindled by the COVID-19 pandemic \cite{Lallo2021}. This
includes ethical concerns regarding the accountability of these intelligent
systems, clear definitions of levels of autonomy \cite{Yang2017},
and so on.

In line with the views of other groups regarding the next frontiers
in robotics \cite{S_nderhauf_2018,Lallo2021}, we see a future in
which AI-robots conduct experiments in challenging environments while
interacting with scientists as peer scientists.\footnote{https://sites.google.com/g.ecc.u-tokyo.ac.jp/moonshot-ai-science-robot/}
This is part of the ambitious Moonshot Research \& Development Program
of the Cabinet Office of Japan\footnote{https://www8.cao.go.jp/cstp/english/moonshot/top.html}
to have, by 2050 and beyond, AI-robots that autonomously learn, adapt
to their environment, evolve in intelligence, and act alongside human
beings.

To achieve that long-term goal, our project is divided into many fronts.
In one of those, discussed herein, we are interested in developing
an experimental platform that will serve to conduct robotic experiments
and scientific discovery. Indeed, one can think of the AI as the \emph{brain},
whereas the robotic platform is the \emph{body}. In this work, we
present our first iteration of the \emph{body} of the AI-robot science
platform we envision together with elements that make one layer of
the \emph{brain}, i.e. the kinematic-level control algorithms and
human-machine interfaces. The multi-arm platform has several novel
elements regarding hardware and software. We also developed two matching
digital twins, which we make openly available.

Our novel robotic platform has been designed to study autonomous scientific
experiments on animals and plants. Unlike artificial objects for which
industrial robots have commonly been used, organisms have considerable
individual variation and heterogeneity of mechanical properties. This
often implies that the experiments are composed of unpredictable and
unique elements necessitating manual operation by skilled operators,
and consequently being difficult to automate. This exposes personnel
to health risks, such as infectious diseases. Therefore, one can expect
that automation would enable safer applications and potentialize other
challenging tasks in unwelcoming environments, e.g. space.

In this context, the manipulation targets can vary between several
hundred micrometers to several millimeters in size. Because they often
require high magnification for observation, the field of view is extremely
narrow and conventional surgical robotic technologies cannot be applied
as-is. In addition, conventional automation has often been studied
with the aim of reproducing human movements. However, there must be
a robotic configuration that is easier for a robot system to conduct
the task even if the configuration is difficult for a human to teleoperate.
Therefore, in our prototype, the manipulator robots' bases are placed
on a circular rail, to allow each robot arm to autonomously position
its base as part of the kinematic structure.

In effect, we developed a platform for research in which the robot
system can be used to learn skills through the teleoperation performed
by human operator(s) and gradually evolve into to a more robot-centric
approach, increasing its degree of autonomy. We do not believe that
the robot's trajectory itself is directly related to skills, but that
skills can be found in the interaction between the robot and the object.
Thus, the robots do not need to move in a similar way to humans, as
long as the task is completed with enough quality. The proposed system
is equipped with four arms with different characteristics; the choice
of the number of arms is tentative, and a main goal is for the system
to have a sense of ``self''. In the future, we expect it to be able
to autonomously choose the best combination of robot arms and tools
for a given task. In this way, the platform has been designed to be
scalable to accelerate research into autonomous non-repetitive tasks
on small natural objects.

In order to have a concrete aim, our initial prototype has been inspired
by scientific discovery related to \emph{intravital} \emph{imaging}
\cite{Pittet_2011}\emph{.} Intravital imaging refers to images taken
while the organism is alive, through a type of window. One of the
applications of this imagining technique is to study the growth of
human cells \cite{Takebe2013}. This paper reports that the platform
is now able to remotely operate on minute objects and collect training
datasets to study skills, while automatically avoiding collisions
outside of its extremely narrow field of view. As a proof-of-concept,
we perform experiments under teleoperation, namely a Tokyo\textendash Kyoto
teleoperated mock experiment using eggs during the IROS'22 Exhibition,
peg transfer, gauze cutting, and the world's first four-hand teleoperated
drilling for a cranial window, by users with no prior surgical experience.

\section{Related works}

In recent years, teleoperated robotic platforms have gained the scientific
community's attention thanks to the emergence of new applications
in several fields, especially in surgical robotics \cite{dupont_decade_2021},
where surgeons exploit the robot's dexterity to perform challenging
medical procedures. 

Despite the high level of accuracy of the robotic platforms, the quality
of teleoperated robot-assisted surgery (RAS) often relies on the surgeon's
skills and experience, which are subject to human imprecision \cite{saeidi_autonomous_2022}.

Aiming for greater robot autonomy, several platforms have been developed
for surgical robotics research, for instance the dVRK \cite{davinci_research_kit_2014}
and the RAVEN \cite{hannaford_raven-ii_2013}. The former is based
on the first-generation da Vinci robot hardware and the latter is
based on custom open hardware. Both systems are based on cable-driven
mechanisms, and, different from serial manipulators, they often require
additional strategies to compensate for the inaccuracies related to
slack and stretch of the cables \cite{peng_real-time_2020,hwang_efficiently_2020}.
Because of that, several platforms based on industrial serial manipulators
have been proposed \cite{mayer_endopar_2004,hagn_dlr_2008,Marinho2020}.

Schwaner et al. \cite{schwaner_mops_2021} presented MOPS, an open
platform for surgical robotics research based on serial manipulators.
MOPS has high modularity, uses commercially available robot arms,
and allows scalability with custom tools. In addition to all these
features, our platform includes a novel hardware design, which is
oriented to dexterous cooperative task manipulations, and targets
general experiments for scientific exploration.

\section{Robotic system overview}

\begin{figure*}[t]
\begin{centering}
\includegraphics[width=1\textwidth]{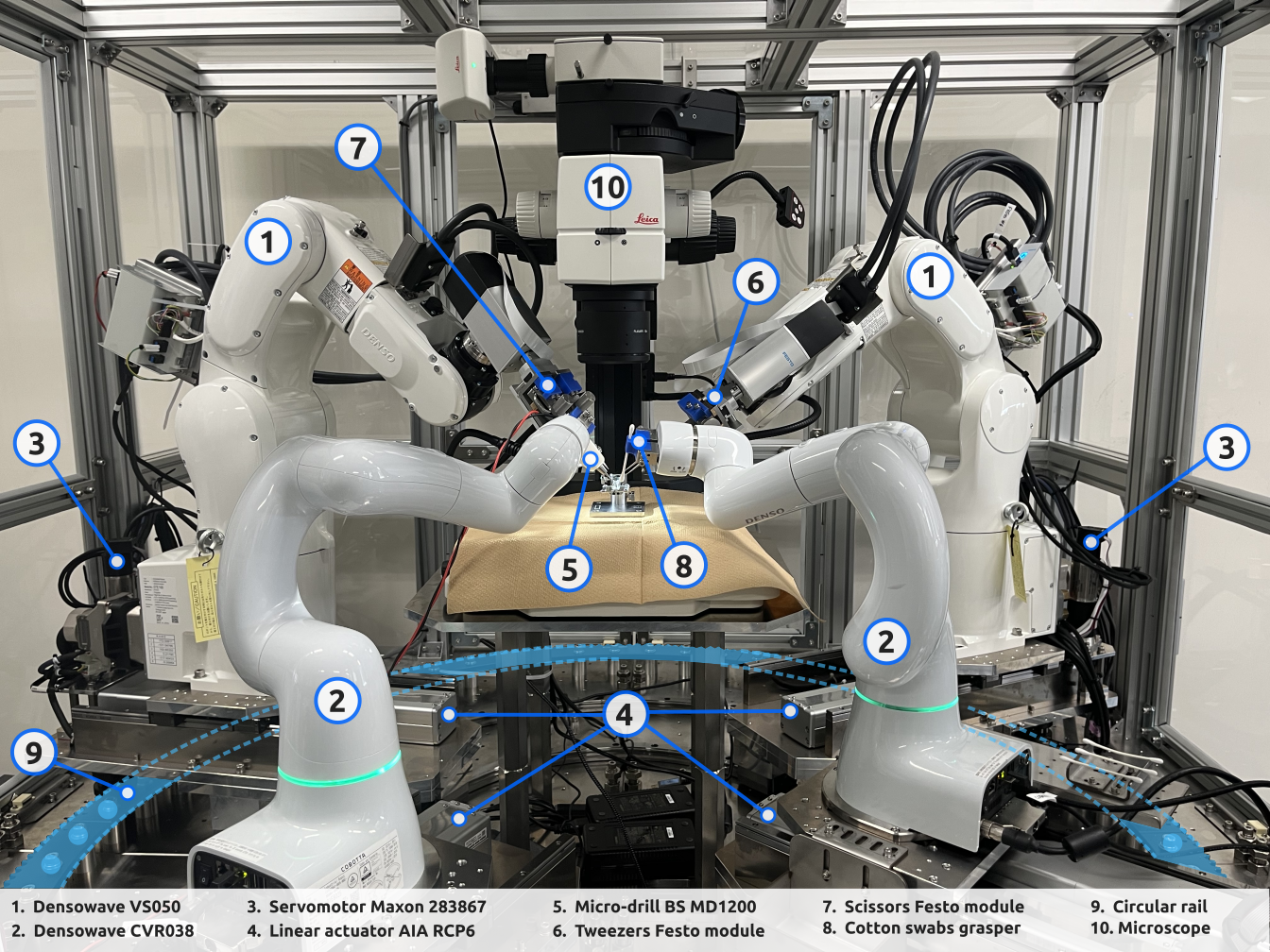}
\par\end{centering}
\caption{\label{fig:AISP}Our multi-arm robot platform. The four branches can
be individually controlled. Two branches contain a collaborative-type
manipulator robot and two branches contain an industrial-type manipulator
robot. Each manipulator robot's base is attached to a linear actuator
which is, then, attached to the circular rail using individual motors.
The system has, in total, 34 active degrees-of-freedom in the configuration
space, in addition to two degrees-of-freedom for actuating the tweezers
and scissors.}
\end{figure*}

The proposed robotic system is shown in Fig. \ref{fig:AISP}. The
whole system is enclosed by a frame. Our proposed novel design is
based on a circular rail to which each robotic branch is attached.
Each branch has a servomotor (283867, MAXON, Switzerland) to independently
rotate about the rail. There is no mechanical limitation for the motion
of each branch about the rail.\footnote{In effect, the rotation of each branch about the central reference
frame is limited by the length of the cables connecting each branch.} Serially, each branch has a linear actuator that moves back-and-forth
in the direction of the center of the rail (RCP6, AIA, Japan).

Composed with this novel multi-arm mechanism, we have four robot branches.
Two branches are composed by a collaborative-type robot (CVR038, Densowave,
Japan) attached serially. The remaining two branches are composed
by the serial connection of an industrial-type robot (VS050, Densowave,
Japan). By doing so, two robotic branches have 8 degrees of freedom
and the remaining two branches have 9 degrees of freedom.

Aiming at a large set of possible applications including the cranial
window, we have four customized end-effectors\footnote{More information available in the supplementary material.}.
One is a customized micro-drill based on a commercial micro-drill
(MD1200, Braintree Scientific, USA). Another is a customized grasper
for cotton swabs. The last two are customized actuators based on a
rotary gripper module (EHMD-40-RE-GE, FESTO, Germany) to operate tweezers
(15-416, BRC, Japan) and scissors (S7, Tsubasa Kougyo, Japan) that
are commonly used in cranial window procedures.

Vision of the workspace is provided through six 4K cameras (STC-HD853HDMI,
Omron-Sentech, Japan).\footnote{All six cameras are mounted in fixed custom positions and orientations
according to the task requirements.} Four of these cameras are placed around the workspace directed at
the center to provide an overall understanding of the state of the
robotic system using wide field-of-view lenses (VS-LDA4, VS Technology,
Japan). Another camera is used to provide vision for the cranial window
task using small field-of-view lenses (VS-LDA75, VS Technology, Japan).
The last camera provides a more balanced field-of-view, for tasks
such as those included in the fundamentals of laparoscopic surgery\footnote{\url{https://www.flsprogram.org}}
(FLS).

\subsection{The fundamentals of laparoscopic surgery}

The fundamentals of laparoscopic surgery (FLS) is a program to educate
and assess surgeons in laparoscopy. One part of it comprises a training/assessment
kit containing a number of tasks that aim to teach surgeons how to
perform surgical procedures using long-thin instruments, through a
small incision (emulated by a box or plate with small holes), and
under indirect view of a laparoscopic camera. Among the existing tasks,
peg-transfer involves picking up blocks and transferring them between
pegs. It is important to note that, such task, \emph{per se}, has
no medical significance. Instead, it aims to train the surgeons' dexterity
by abstracting away the task. In addition, it helps the surgeons learn
the hand-eye coordination and task constraints. The FLS tasks are
commonly used as a benchmark for robot dexterity and for the evaluation
of AI algorithms. In robotics research, the peg transfer is still
a benchmark for robot dexterity and AI research \cite{Hwang2023},
even though it is one of the easiest tasks for humans given that the
blocks are rigid and easy to visually discern. More complicated tasks,
such as gauze cutting \cite{Thananjeyan2017}, are still an open problem
in robotics research. It involves using one of the robotic arms to
create tension in the gauze while another arm cuts it. Many groups
are exploring the use of state-of-the-art simulators to address this
challenge, by trying to learn flexible material dynamics \cite{Ou2023}.
Other tasks involve handling sutures, performing knot tying, and handling
suture threads with varying degrees of difficulty.

\subsection{Design novelty}

The proposed platform is the culmination of a longstanding know-how
in our group in dexterous manipulation in constrained workspaces,
starting from mechanical systems developed for a particular purpose,
e.g. \cite{Mitsuishi2012}. Aiming to cover a larger number of possible
applications, we have moved on to using general-purpose manipulators
with specialized end-effectors \cite{Marinho2020}, effectively targeting
a larger number of applications with arbitrary constraints \cite{Marinho2019}.

With dual arm systems with a movable\footnote{We say movable in the sense that the placement of the robot can be
changed by a person (e.g. by unlocking and pushing the cart in which
the robot is placed), but the base placement is not an actuated degree-of-freedom.} base (e.g. on a cart with casters), we have identified two challenges,
which also affect similar robotic systems. These inspired the rail-plus-linear-actuator
design. First, that for model-based (self) collision avoidance, i.e.
the sense of ``self'', the precise knowledge of the base position
of each manipulator is paramount. In this context, a time-consuming
high-accuracy (re-)calibration for each new base repositioning is
a bottleneck for the research. Second, that when arms with redundancy,
e.g. each SmartArm with 9 degrees of freedom \cite{Marinho2019},
workspace optimization using a performance metric, e.g. manipulability,
is an open challenge. The redundancy, complex nature of the tasks,
and large number of constraints, makes it impossible to optimize the
placement of the bases of the robots using existing techniques. Having
the rail-plus-linear-actuator, the dexterous workspace of the manipulators
becomes its original volume, extruded by the area of the rail-plus-linear-actuator,
effectively solving that issue for the table centered in the platform.
In addition, because all those additional degrees of freedom have
an encoder, we do not need re-calibration after motion. It is important
to note that, given that there is considerable overlap on the workspaces
of all arms, our design is only feasible given the existence of state-of-the-art
execution-time (self) collision-avoidance frameworks, e.g. \cite{Marinho2019}.

In addition, the system being composed by two different types of arms
is inspired by applications in which the larger robotic arms cannot
be used, owing to, e.g., size restrictions \cite{Buendia2019a}. In
such applications, the large payload of larger arms is unnecessary
and their large volume becomes, instead, an issue. They can easily
occlude the view of the target and require that the camera is set
considerably farther from the manipulation target, further restricting
the choice of suitable camera/lens systems.

A multi-arm system, such as the one described herein, addresses all
these issues and enables the scientific exploration of a large number
of applications without a mechanical change. Having such a structure
allows us to learn tasks in this system and, when possible, transfer
the learned aspects to systems with a lower number of arms.

\section{System integration \& software}

\begin{figure*}[t]
\begin{centering}
\includegraphics[width=1.6\columnwidth]{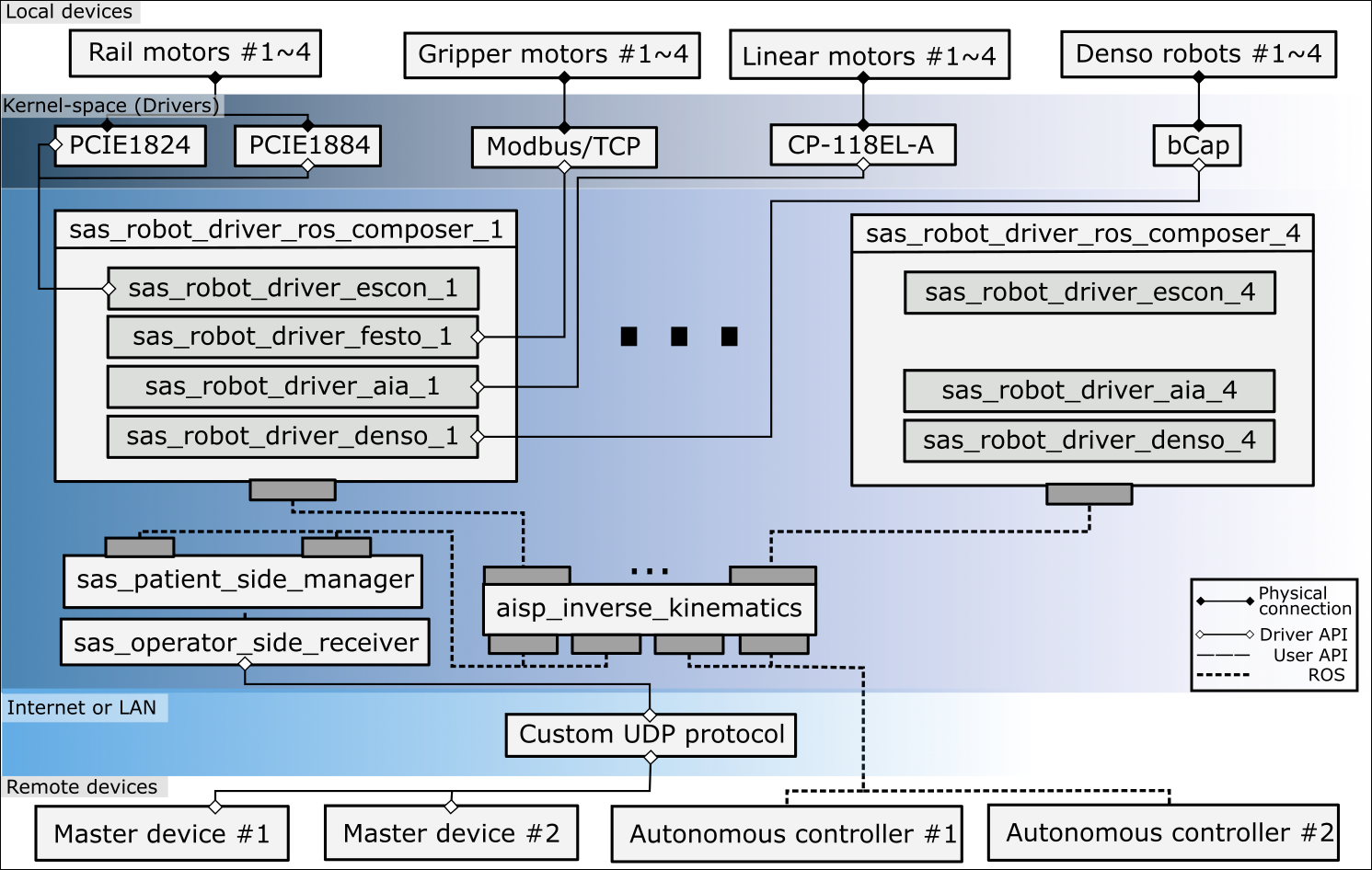}
\par\end{centering}
\caption{\label{fig:ros_system_integration}An example of the system architecture
when two robots are teleoperated and two robots are autonomously controlled.
The local devices are controlled from the centralized control computer.
Each robotic branch is encapsulated in a \texttt{sas\_robot\_driver\_ros\_composer}.
There are four branches, but only two are shown in the figure for
simplicity. Those four robotic branches are controlled in joint-space
by the \texttt{aisp\_inverse\_kinematics} node which takes into account
joint limits, environmental constraints, and collision avoidance among
robots. The aisp\_inverse\_kinematics node receives desired pose signals
for each branch individually. The desired pose signals can be sent
through a master device, a custom program, or a ROS-based program.}
\end{figure*}

In the implementation described herein, we used general-purpose computational
platforms to make the system feasible for experiments in scientific
exploration. Our target at this stage was to develop a system with
acceptable latency to a human operator or an autonomous system with
a similar level of performance. In this context, we did not focus
on developing a system with hard real-time guarantees, hence we have
no strict guarantee of adherence to a predefined periodicity in any
given computational thread. We took reasonable steps to reduce lag
up to the reported level of responsiveness, but a missed thread deadline
was not catastrophic in our target applications. If a system inspired
by ours is to be used in a critical application that requires such
guarantees, a suitable real-time kernel (e.g. RT\_PREEMPT\footnote{\url{https://wiki.linuxfoundation.org/realtime/start}}
used in \cite{Marinho2020}) along with the real-time capabilities
provided by a suitable robotic system middleware\footnote{Hard real-time was intractable in ROS 1, but is available in ROS 2:
\url{https://docs.ros.org/en/humble/Tutorials/Demos/Real-Time-Programming.html}.
Forks such as Space ROS might also be of interest \url{https://space.ros.org/}.} might be good starting points. Nonetheless, such adjustment is not
expected to be trivial.

Our system has been designed to have flexibility on the control modes.
This means that any of the four robots can be controlled through teleoperation
or autonomously. Each robot can be independently controlled by its
respective operator side, which can be anywhere in the world with
a properly configured internet connection. In this light, our system
can be divided into \textcolor{black}{multiple} \emph{operator sides}
(OS) and a \emph{follower side} (FS)\@. One example configuration
is shown in Fig.~\ref{fig:ros_system_integration}.

\subsection{Operator sides}

The OS are enabled by the \texttt{SmartArmMaster} program, developed
as part of this work and made available with an installer\footnote{\url{https://github.com/SmartArmStack/smart_arm_master_windows/releases/latest}}
on Windows,\footnote{Owing to limitations on the current version of the manufacturer's
drivers, reliable support for two masters on the same computer was
only available on Windows.} free for noncommercial use. It is currently compatible with 3D Systems'
haptic devices.\footnote{\url{https://www.3dsystems.com/haptics}}
We developed a similar program for Medicaroid's hinotori's master
manipulator device,\footnote{\url{https://www.medicaroid.com/en/product/hinotori/}}
used in Section~\ref{subsec:Tokyo=002013Kyoto-Teleoperated-Eggshel}.
Our master programs read the kinematic information of the devices
through the vendor's API, encode that information into a binary stream,
and send it to the FS through user datagram protocol (UDP).

\subsection{Follower side}

Regarding the FS, the system's information is processed through two
computers. The first computer, based on Windows 10 x64, is responsible
for all tasks requiring GPU support including running the digital
twins\footnote{During our development, the reliability of IsaacSim on Windows was
considerably better than on Ubuntu.} described in Section~\ref{sec:Digital-Twins}. In addition, it is
used to handle and capture the information from the cameras and generate
the composed views used in the experiments described in Section~\ref{sec:Experimental-showcase}.
The second computer, based on Ubuntu 20.04 x64 and ROS Noetic\footnote{An Ubuntu 22.04 x64 with ROS 2 Humble version is also available.},
is responsible for monitoring and controlling all robotic devices
in a centralized manner.

\subsubsection{Low-level control}

The low-level control pertains commanding the controllers of the four
rail motors (ESCON 50/5, MAXON, Switzerland), the four motors in total
for the two rotary gripper modules  (CMMO-ST-C5-1-LKP, FESTO, Germany),
and the four linear actuators (PCON-CB, AIA, Japan). The manipulator
robots have the same type of controller (RC8, DensoWave, Japan) that
is controlled over bCAP Slave Mode\footnote{\url{https://www.denso-wave.com/en/robot/product/function/b-CAP.html}},
through UDP.

This integration has been achieved through the development of a large
number of custom ROS Noetic packages,\footnote{A ROS 2 version is also available.}
all free for noncommercial use.\footnote{\url{https://github.com/SmartArmStack}}
Considering low-level control, the \texttt{sas\_robot\_driver\_denso}
and \texttt{sas\_robot\_driver\_escon} were re-designed in this work,
based on \cite{Marinho2020}. The software packages \texttt{sas\_robot\_driver\_festo}
and \texttt{sas\_robot\_driver\_aia} were newly developed using pymodbus.\footnote{\url{https://github.com/riptideio/pymodbus}}

\subsubsection{Composition of robots}

The composition of several devices\footnote{More information available in the supplementary material.}
into a single serial robotic system is enabled by an instance of \texttt{sas\_robot\_driver\_ros\_composer}.
Using a configuration file, we specify which \texttt{sas\_robot\_driver\_ros}
are to be composed and in what order. In addition, we can add CoppeliaSim
information for each composed robot. The abstraction of any number
of systems into a single serial robot facilitates and modularizes
the higher-level control modules which is impervious to changes in
the lower-level devices. In this implementation, we have a total of
four branches, each abstracted by a \texttt{sas\_robot\_driver\_ros\_composer}.

\subsubsection{Centralized control\label{subsec:Centralized-control}}

The centralized control module is the main contribution in terms of
software. It is the implementation of a centralized kinematic control
strategy based on quadratic optimization and inequality constraints
\cite{Marinho2019a,Marinho2019}. The centralized control can be configured
to simultaneously handle any number of robot branches, in this case
four, as follows
\begin{equation}
\argminimtwo{\myvec u\in}{\dot{\myvec q}}{\sum_{i=1}^{4}\mathcal{F}_{i}\left(\myvec q_{i},\myvec x_{d,i}\right)}{\myvec W_{s}\dot{\myvec q}\preccurlyeq\myvec w_{s}}{\mymatrix W_{p}\dot{\myvec q}\preccurlyeq\myvec w_{p}},\label{eq:QP_problem}
\end{equation}
where $\mathcal{F}_{i}\left(\myvec q_{i},\myvec x_{d,i}\right)$ is
a convex objective function related to the task of each robotic branch
\cite{Marinho2019a}, $\myvec q_{i}$ are the joint values of the
$i$th robot, $\myvec q$ is the vector stacking all $\myvec q_{i}$
in order, $\myvec x_{d,i}$ is the task-space target for the $i$th
robot, and $\myvec u$ is the vector of joint velocities output by
the solver that will be send to the robotic system.

With respect to the constraints, the first set of constraints contains
the block-diagonal matrix $\mymatrix W_{s}$ that encodes constraints
related to a \emph{single} robot branch. Those can be, for instance,
joint limits, joint velocity limits, and collision avoidance with
static objects in the workspace using vector-field inequalities (VFIs)
\cite{Marinho2019}. Lastly, the second set of constraints is related
to \emph{pairwise} constraints to prevent collisions between robotic
branches.

The inequality constraints are used to provide (self) collision avoidance,
that is, a sense of self-awareness to the robotic system. That is
very important in such a complex system, because a naive control strategy
certainly would cause the robot to break. The self-awareness is provided
through our VFI methodology. Each VFI is based on the signed distance
between geometric primitives.

One of the biggest challenges we had so far with our strategy was
related to the cumbersome process of adding several primitive pairs.
In this work, we partially solve this issue by providing a way to
add each VFI through a configuration file (see Fig.~\ref{fig:example_of_two_vfi_configs}).\footnote{A minimal example using ROS and ROS2 is available in https://github.com/AISciencePlatform/aisp\_ros\_control\_template
and https://github.com/AISciencePlatform/aisp\_ros2\_control\_template,
respectively.}

\begin{figure}[tbh]
\begin{centering}
\begin{lstlisting}[breaklines=false,captionpos=b,frame=tb,keywordstyle={\color{blue}},alsoletter={*()"'0123456789.},alsoother={\{\=\}},backgroundcolor={\color{lightgray3}},basicstyle={\footnotesize\ttfamily},breaklines=true,commentstyle={\itshape\color{lightgray}},fillcolor={\color{lightgray3}},framexleftmargin=1em,framextopmargin=1em,keywordstyle={\color{magenta}\bfseries},language=Python,literate={{=}{{{\color{blue}=}}}1},morestring={[b]"},stringstyle={\color{blue}\ttfamily},morecomment={[n][\keywordstyle]{{}{}}},morekeywords={vfi_type, cs_entity_environment, cs_entity_robot, entity_environment_primitive_type, entity_robot_primitive_type, robot_index, joint_index, safe_distance, direction},otherkeywords={cs_entity_one, cs_entity_two, entity_one_primitive_type, entity_two_primitive_type, robot_index_one, robot_index_two, joint_index_one, joint_index_two},prebreak={\textbackslash},sensitive=true,stepnumber=1,tabsize=4,upquote=true]
vfi_array:
 -
    vfi_type: "ENVIRONMENT_TO_ROBOT"
    cs_entity_environment: "Cylinder"
    cs_entity_robot: "Denso2_VS050_vfi_sphere_1_0"
    entity_environment_primitive_type: "LINE"
    entity_robot_primitive_type: "POINT"
    robot_index: 3
    joint_index: 1
    safe_distance: 0.180625
    direction: "FORBIDDEN_ZONE"
 -
    vfi_type: "ROBOT_TO_ROBOT"
    cs_entity_one: ["Cobotta1_vfi_sphere_0_0"]
    cs_entity_two: ["Cobotta2_vfi_sphere_0_0"]
    entity_one_primitive_type: "POINT"
    entity_two_primitive_type: "POINT"
    robot_index_one: 1
    robot_index_two: 0
    joint_index_one: 0
    joint_index_two: 0
    safe_distance: 0.16
    direction: "FORBIDDEN_ZONE"
\end{lstlisting}
\par\end{centering}
\caption{\label{fig:example_of_two_vfi_configs}Example of two VFIs described
using a yaml configuration file. The first object describes a forbidden
zone between a point in the fourth robotic branch, related to the
second joint, with respect to a line in the workspace. The second
object describes a forbidden zone between two robots. The first entity
is a sphere related to the second robotic branch's first joint, and
the second entity is related to the first robot branch's first joint.
Using a total of 31 VFIs, we can guarantee that the system is highly
reactive and has collision-avoidance guarantees.}
\end{figure}

\section{Digital Twins\label{sec:Digital-Twins}}

The simulation of our robotic platform is addressed using two different
environments, namely CoppeliaSim,\footnote{https://www.coppeliarobotics.com/}
and Isaac Sim.\footnote{https://developer.nvidia.com/isaac-sim} The
former is mainly used to define geometric primitives, useful to prevent
(self)-collisions by means of the VFIs framework. The latter implements
a digital twin, and it is used to train operators before using the
real platform. Both environments run on the (FS) Windows computer,
which is equipped with an Intel i9-11900K with 64GB RAM and a GPU
Nvidia RTX A6000.

\begin{figure*}
\def\svgwidth{1.0\columnwidth}
\noindent \begin{centering}
\includegraphics[width=2\columnwidth]{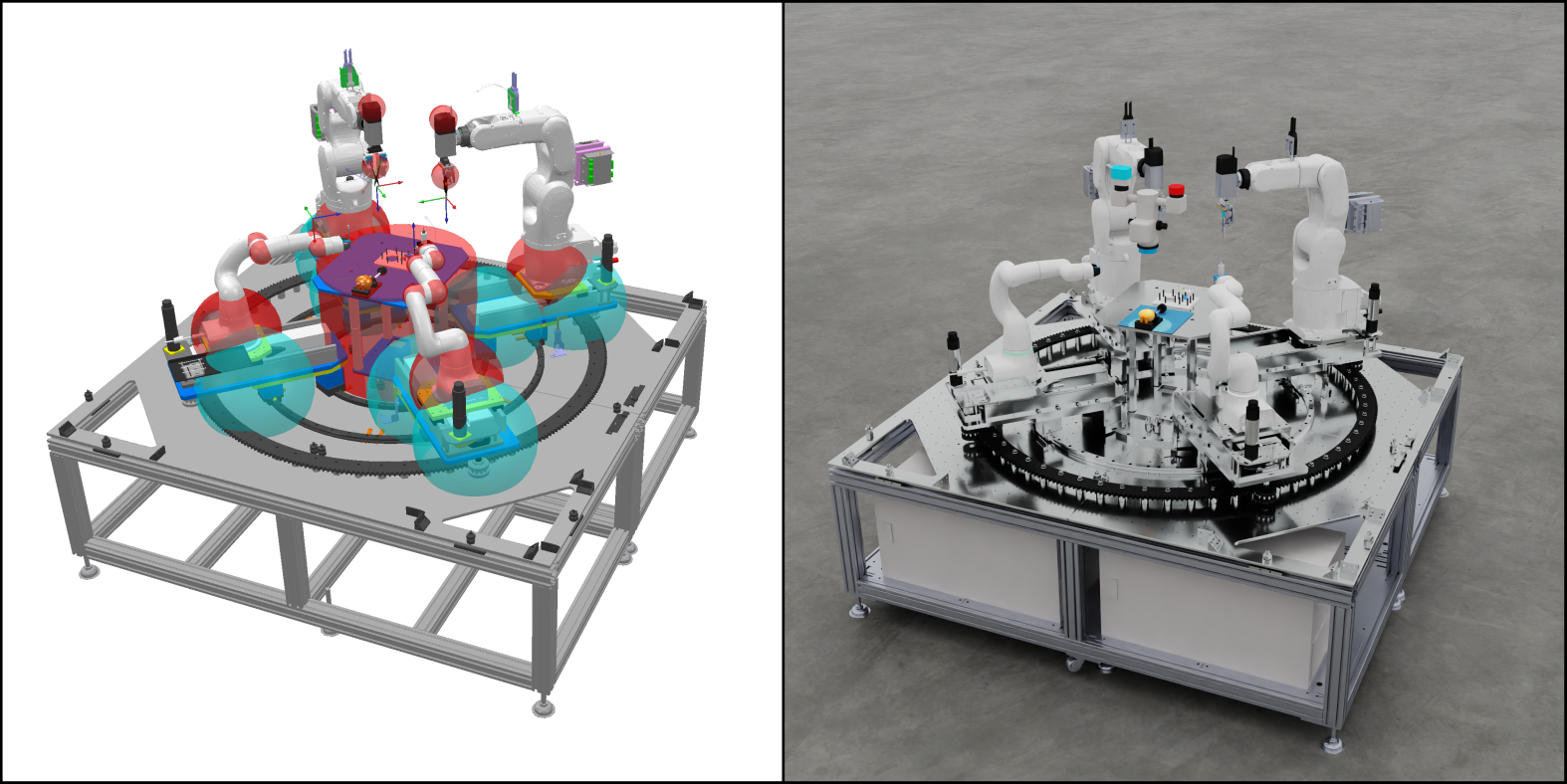}
\par\end{centering}
\caption{On the \emph{left}, a snapshot of the CoppeliaSim scene. The blue
and red spheres are geometric primitives defined by the user. The
kinematic controllers use these primitives to impose VFIs that prevent
(self)-collisions. On the \emph{right}, a snapshot of the Isaac Sim
scene using path tracing render. \label{fig:Path-tracing-render}}
\end{figure*}

\textcolor{black}{The kinematic controllers, which run on Ubuntu and
ROS (FS), receive haptic interface (OP) teleoperation commands and
set the joint positions of the CoppeliaSim simulation. Finally, Isaac
Sim receives the joint position commands from CoppeliaSim by means
of the DQ Robotics library \cite{Adorno-Marinho2020} and the Isaac
Sim Python API (}standalone workflow\textcolor{black}{)}. Fig.~\ref{fig:Path-tracing-render}
shows the render of the scene using the path tracing algorithm.

\section{Experimental showcase\label{sec:Experimental-showcase}}

\subsection{Digital Twin: Isaac Sim}

The proposed digital twin aims mainly to accurately replicate the
robot kinematics, while the robot dynamics needs only to be plausible.
The rationale behind this is twofold. First, the inertial parameters
(e.g., mass and moment of inertia) of the robots composing our system
are not made available by the companies. Furthermore, they are commanded
by joint positions inputs and operate under relatively low velocities
and accelerations in the applications we describe. This allows the
use of kinematic control strategies \cite{Marinho2019}, as in (\ref{eq:QP_problem}),
which do not take into account the robot dynamics. Therefore, to obtain
a similar behavior between the real platform and the Isaac Sim digital
twin,\footnote{More information available in the supplementary material.}
we set the damping and stiffness joint parameters to large values,
and the masses of the links to small values. Both the center of mass
and the inertia matrix of the links are computed automatically by
the simulator according to the collision mesh of the body, which are
approximations of the body mesh by convex decomposition.\footnote{We set the damping and stiffness joint parameters to $10^{6}$ and
$10^{7}$, respectively.}

To validate the proposed approach, we performed two challenging task
simulations. The first one is the peg transfer task, and the second
one is cranial microsurgery on a mouse. Both tasks are performed via
teleoperation. The main goal is to develop a training platform to
improve the operator's skills before using the real robotic system.

\subsubsection*{Peg Transfer Task}

To practice grasping tasks, we implemented a peg transfer task simulation
using the kinematic branch equipped with the forceps. The goal is
to transfer a block from its initial peg to a different one while
avoiding collisions between the robot's forceps and the pegs, as shown
in Fig.~\ref{fig:IsaacSim_examples}. Since the grasping of general
objects is a challenging task, which usually relies on specialized
simulation environments \cite{graspit2004}, we implemented a fake-grasping
strategy. The idea is to constrain the relative pose between the forceps
and the block when they are in contact below a minimum distance. Once
the teleoperator opens the forceps, the block is released.

The block is a hollow triangular prism, which is composed of a rigid
core covered with a deformable mesh, as shown in Fig.~\ref{fig:IsaacSim_examples}.
The former enables static interactions with the forceps, whereas the
latter allows for dynamic interactions with the pegs. Each peg is
linked to a fixed base using a spherical joint set with custom damping
and stiffness properties. This approach allows small movements of
the peg when subjected to external forces, and prevents numerical
instabilities when the robot forceps touch the peg.

\subsubsection*{Cranial Microsurgery}

We implemented a second simulation aiming to practice cranial microsurgery
experiments using the kinematic branch equipped with the drill. Instead
of drilling specific meshes, which is not a trivial solution for Isaac
Sim, we adopted a place-based approach. The goal is to create and
place small spheres on the target object when the contact force applied
by the end-effector exceeds a threshold. By doing so, we include different
behaviors for different force values. This feature allows setting
different colors for the drawn spheres or different textures for the
drilled object, as shown in Fig.~\ref{fig:IsaacSim_examples}.

\subsubsection*{Results and Discussion}

\begin{figure*}
\def\svgwidth{1.0\columnwidth}
\noindent \begin{centering}
\includegraphics[width=2\columnwidth]{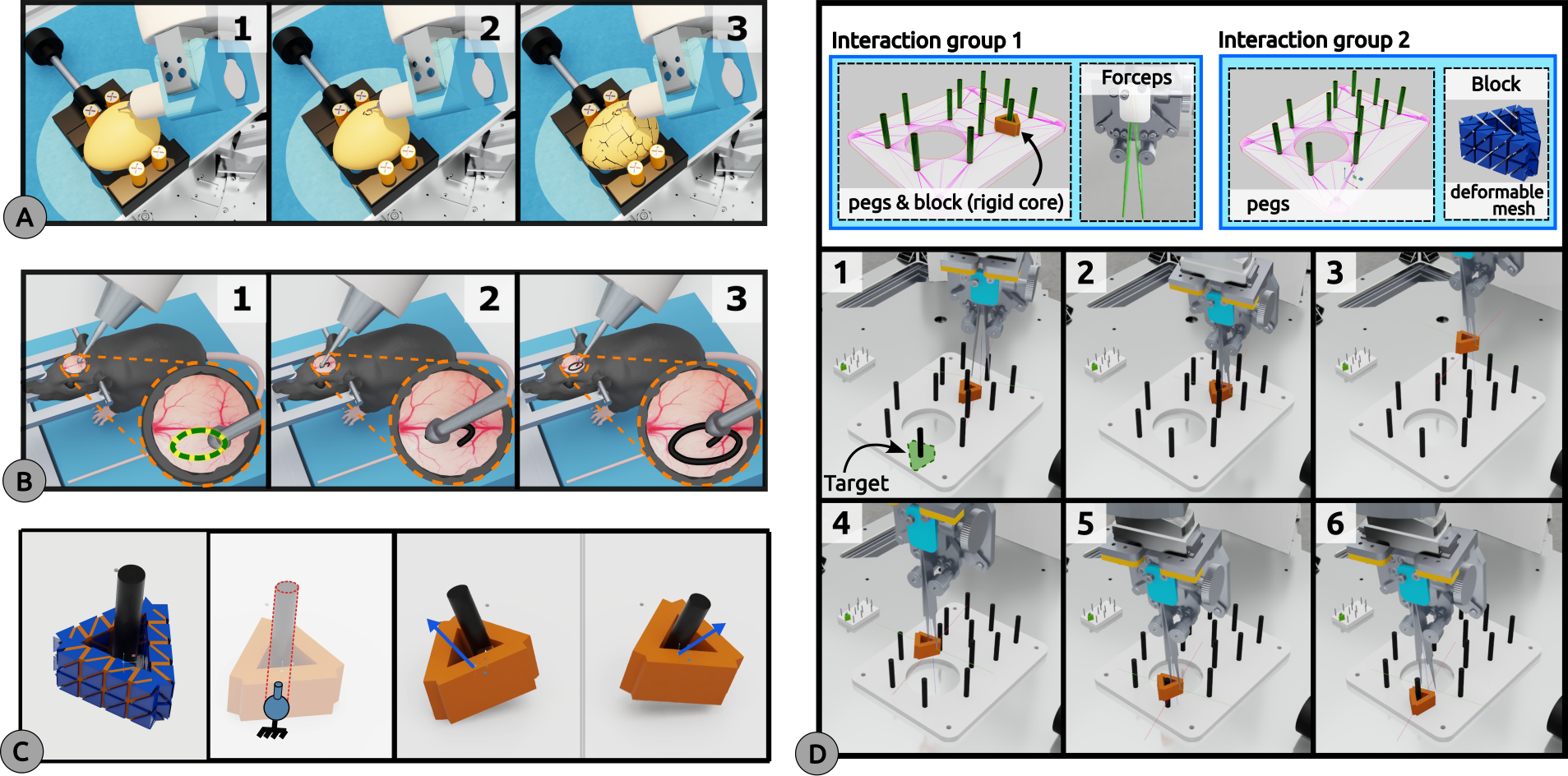}
\par\end{centering}
\caption{$\boldsymbol{A}$: the egg-drilling task. From left to right: The
drill is near the egg, but there is no contact force; the drill applies
to the egg a contact force, and small black spheres are placed on
the egg's surface. The applied force is above a threshold and the
egg texture is modified to imitate a broken egg. $\boldsymbol{B}$:
snapshots of the cranial microsurgery simulation. The desired trajectory
is denoted by the green dashed line. $\boldsymbol{C}$: Interaction
between the block and the peg. From left to right: The block is composed
of a rigid core (orange hollow triangular prism) covered with a deformable
blue mesh. A spherical joint links the peg with a static base; different
behaviors of the block and the peg when subject to an external force.
The blue arrow represents the external force vector applied. $\boldsymbol{D}$:
snapshots of the peg transfer task simulation. The goal is to transfer
the block from its initial peg to a different one while avoiding collisions
between the robot's forceps and the pegs. The collision groups are
denoted on the top. The forceps (including the force sensor) interact
with the rigid core and the pegs. The interactions between the block
and the pegs are handled by the deformable mesh only.\label{fig:IsaacSim_examples}}
\end{figure*}

Fig.~\ref{fig:IsaacSim_examples} shows the snapshots of the peg
transfer simulation task. The block is transferred from its initial
peg to a different one successfully. However, the lack of 3D vision
makes it difficult for the operator to locate the end-effector in
the workspace. To circumvent this, one could add more cameras in different
positions in the scene. Nevertheless, each camera increases the GPU
consumption considerably. Our scene, with one camera, runs at 30 FPS
using a 1920x1080 render resolution. When two cameras are used, the
overall performance drops to about 18 FPS. Another solution could
be the use of a VR headset, which is currently supported by Isaac
Sim. However, that is out of the scope of this work.

Figure~\ref{fig:IsaacSim_examples} shows the snapshots of the cranial
microsurgery simulation task. The goal is to perform an oval trajectory
on the skull surface of a mouse. This trajectory represents a small
piece of skull tissue that is to be removed. In this case, the robot
is not teleoperated. Instead, we define a position task to follow
a closed trajectory. The height of the end-effector is independently
controlled to keep a constant force applied on the skull. The task
is performed successfully, as shown in Fig.~\ref{fig:IsaacSim_examples}.
However, our current simulation does not take into account the deformable
behavior of the mouse skull, which is a challenging feature in the
real experiments.

\subsection{Physical Platform}

\subsubsection{Tokyo\textendash Kyoto Teleoperated Eggshell drilling\label{subsec:Tokyo=002013Kyoto-Teleoperated-Eggshel}}

\begin{figure}[h]
\begin{centering}
\includegraphics[width=1\columnwidth]{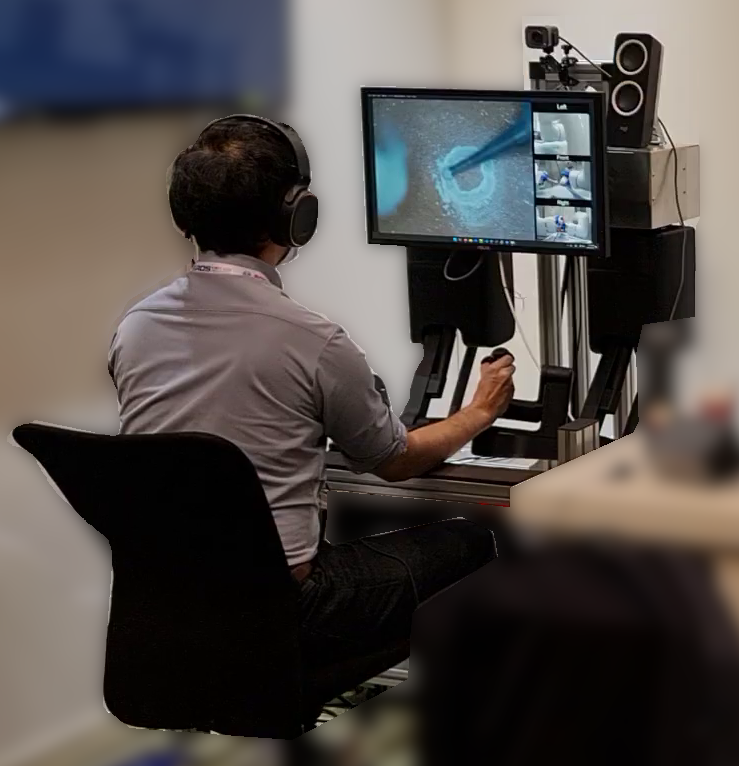}
\par\end{centering}
\caption{\label{fig:Kyoto-side-operator-console.}Kyoto-side operator console.
The operator had access to two control interfaces similar to those
used by the Hinotori surgical robot system. The user had a view composed
of the Tokyo-side microscopic image taking most of the screen and
three smaller views on the right corresponding to the left, front,
and right views of the Tokyo-side system. In addition, the operator
could hear the drilling sounds using speakers.}
\end{figure}

One of the first and most delicate tasks in a cranial window is to
drill a 8 mm patch of the mouse skull without damaging the brain below.
Such a task is challenging even for expert scientists, given that
even light damage to the brain results in complete failure.

Before blindly experimenting on mice, scientists train using raw chicken
eggs. Chicken eggs have a similar thickness to the mouse skull and
an underlying membrane. The task then becomes to drill and remove
a circular patch of the eggshell without damaging the membrane below.

In this work, to show the reliability of our robotic setup, we designed
a Tokyo\textendash Kyoto experiment as part of an exhibition at IROS'22.
The robotic system was in Tokyo while we had in Kyoto two Hinotori-type\footnote{These master devices were borrowed from Kawasaki heavy industries
as part of a cooperative work to make them available to research users.} master devices, as shown in Fig.~\ref{subsec:Tokyo=002013Kyoto-Teleoperated-Eggshel}.\footnote{We used a 3:1 motion scaling in all experiments.}
The control signals between both sides were sent through suitable
UDP protocols.\footnote{The video was compressed using ffmpeg (https://ffmpeg.org) and sent
through session description protocol (SDP) over UDP, whereas the audio
was streamed using SonoBus (https://www.sonobus.net/).}

\begin{figure}[h]
\begin{centering}
\includegraphics[width=1\columnwidth]{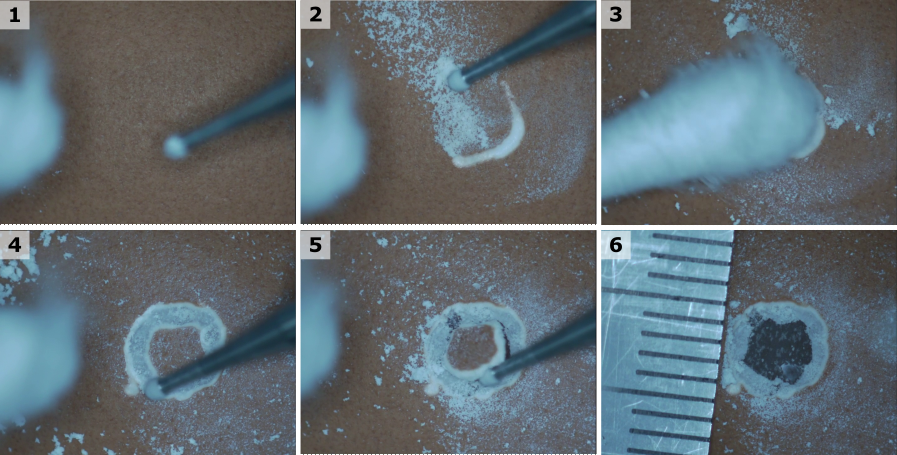}
\par\end{centering}
\caption{\label{fig:A-sample-of}A sample of what could be achieved in the
Tokyo\textendash Kyoto teleoperation experiment. With the right hand,
the user carefully drilled the eggshell. With the left hand, the user
could use a cotton swab to clear the eggshell dust. After about five
minutes of drilling and cleaning, the shell was penetrated without
damaging the underlying membrane. The task was, thus, successful.}
\end{figure}

As a showcase of what could be achieved, we performed one complete
eggshell-drilling task\footnote{https://youtu.be/3JPSywPAdj4}, with
snapshots shown in Fig.~\ref{fig:A-sample-of}.

\subsubsection{Peg transfer}

\begin{figure}[h]
\begin{centering}
\includegraphics[width=1\columnwidth]{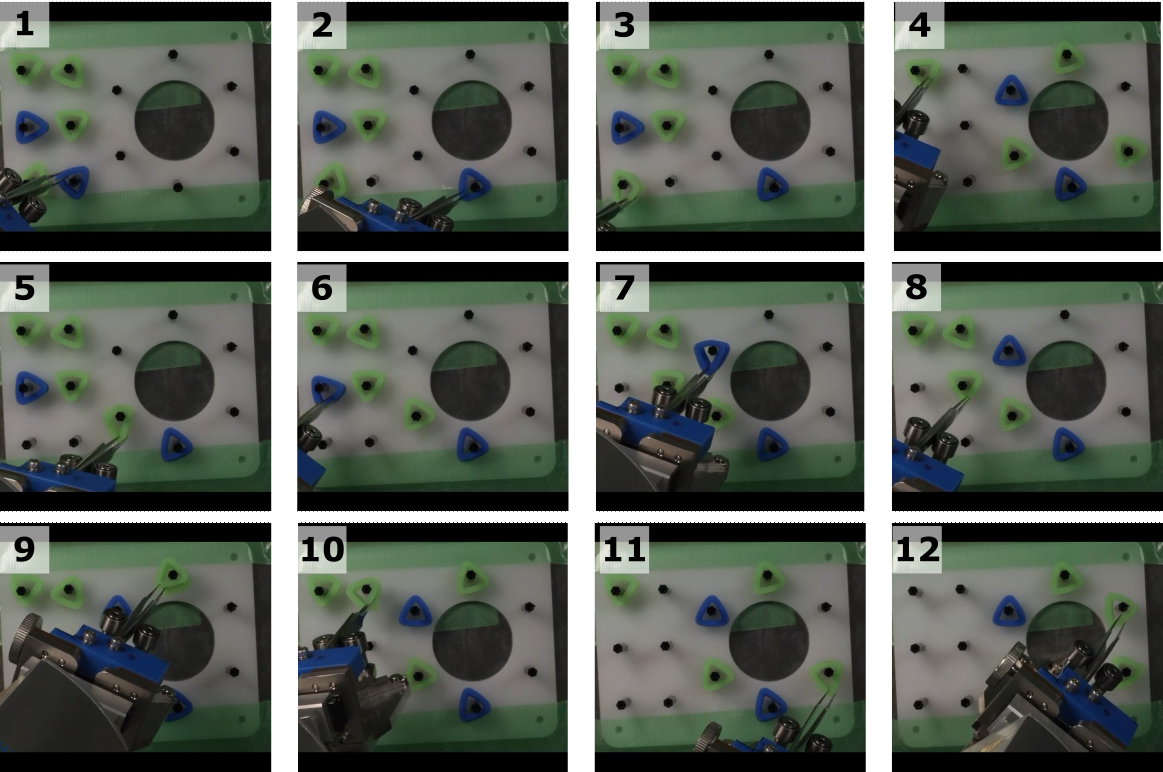}
\par\end{centering}
\caption{\label{fig:peg_transfer}Peg transfer on the real system operated
in teleoperation mode. The peg transfer experiment is ubiquitous in
surgical robotics automation and a common benchmark for human and
robot dexterity. This feasibility experiment serves to show that our
system is capable of both incredibly fine tasks and dexterous tasks
such as peg transfer.}
\end{figure}

The peg transfer task, part of the fundamentals of laparoscopic surgery
(FLS) certification, is ubiquitous in surgical robotics automation
and a common benchmark for human and robot dexterity. In this showcase
experiment, our intention is to show the feasibility of using the
robotic arm with tweezers to perform the peg transfer. The procedure
was performed through teleoperation using the large field-of-view
lenses on top and left, front, and right cameras.

With only the top view, it was extremely challenging to properly perform
the task because of the difficulties in perceiving depth from a single
camera. With the many simultaneous views, the user can naturally compensate
for that difficulty. In this context, the user was able to perform
the task without difficulties as shown in Fig.~\ref{fig:peg_transfer}.
As a reference, the task was done within 3 minutes.

Depending on usage, the system could also be retrofitted with two
tweezer actuation units, allowing for two-arm peg transfer execution.

\subsubsection{Gauze cutting}

\begin{figure}[h]
\begin{centering}
\includegraphics[width=1\columnwidth]{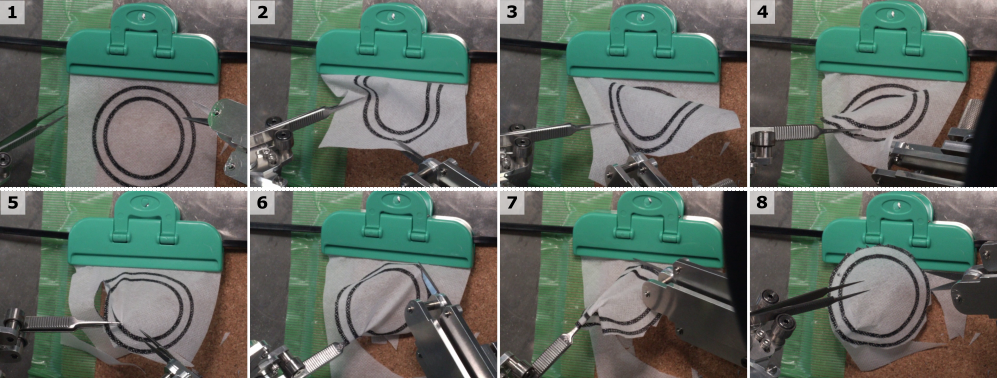}
\par\end{centering}
\caption{\label{fig:gauze_cutting}Gauze cutting using the real system operated
in teleoperation mode. The gauze-cutting experiment is a considerably
challenging task included in the FLS. This experiment serves to show
that the tweezers\textendash scissors pair can be used with high dexterity.}
\end{figure}

The gauze-cutting task, also part of the FLS certification, is an
extremely challenging task for automation owing to the complexity
of interacting with the gauze. In this experiment, our purpose was
to show that the scissors attachment works properly in a mechanical
aspect. In addition, to know if a complex object can be cut by our
integrated system. The cameras used in this experiment were the same
used for the peg transfer, namely: top and left, front, and right
cameras.

This experiment was considerably more challenging than the peg transfer
and snapshots of the gauze cutting are shown in Fig.~\ref{fig:gauze_cutting}.
In particular, a considerable amount of rotation change is required
to allow the task to be feasible. This causes the manipulator robots
to perform, sometimes, large motions in the workspace. Despite that,
the robots can safely move owing to (self) collision avoidance described
in Section~\ref{subsec:Centralized-control}.

\subsubsection{Two-person teleoperated cranial window drilling}

\begin{figure*}[t]
\begin{centering}
\includegraphics[width=2\columnwidth]{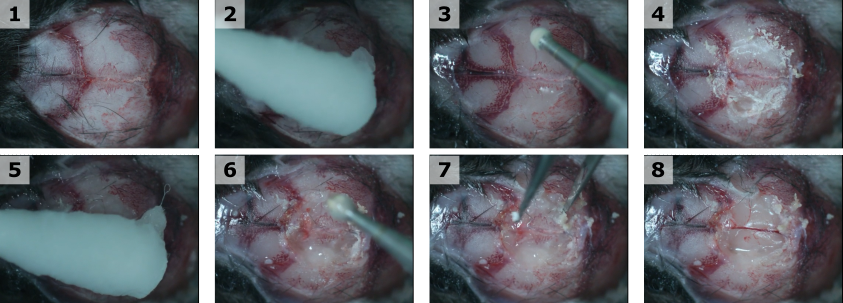}
\par\end{centering}
\caption{\label{fig:teleoperated_cranial_window}The world's first reported
teleoperated cranial window in an euthanized mouse. Two users and
three tools were used. The first user handled the microdrill and the
cotton swab. The second user handled the tweezers that were used to
remove the circular cranium patch. The only feedback available to
the users are the camera images and the sound. This was a feasibility
experiment in which time was of no particular essence; but, nonetheless,
we were able to make a proper circular hole within 6 minutes.}
\end{figure*}
In this experiment, our objective was to use the robotic platform
for drilling a 8 mm patch on the skull of an euthanized mouse without
damaging the tissue below. The mouse used in this experiment was otherwise
going to be discarded and was provided by a neighboring research laboratory,
that is, it was \emph{not} euthanized for this experiment.

The preparation of the mouse for the experiments was done manually
by removing the skin on top of the cranium and fixating the mouse
under camera vision. One operator was tasked with handling the microdrill
and the cotton swab. When the mouse is alive, the cotton swab and
other tools are used to control and stop bleeding. In our case, the
cotton swab was used to keep the cranium wet using tap water inside
a small plastic container. The user operated the robot to wet the
cotton swab and then applied it to the cranial surface. After that,
the user performed the drilling of the skull. Those steps were repeated
for about 5 minutes. After that, the second user, seated in another
computer and having access to their own master interface, operated
the tweezers. The operator could successfully remove the cranial patch.
Snapshots of this procedure are shown in Fig.~\ref{fig:teleoperated_cranial_window}.

\section*{Conclusions}

In this work, we have introduced a multi-arm robot platform for scientific
exploration. We described its design, based on a centralized rail
system and customized end effectors. After showing the scalable system
integration, we showed two digital twins. One digital twin is based
on CoppeliaSim and ready to run on most computers. Another is based
on Isaac Sim and, while more computationally demanding, provides more
realistic graphics and physical accuracy.

To evaluate this platform, we have shown experiments involving the
real platform and the digital twins. We have performed peg transfer,
gauze cutting, mock experiments using egg models, and the drilling
toward cranial window generation using three robot arms controlled
by two operators. The digital twins and related control software are
available free for noncommercial use and can be used by anyone interested
in validating their algorithms in a state-of-the-art multi-arm platform.

There is ongoing work in using this platform for artificial intelligence
research toward scientific exploration. The teleoperation of the system
is paramount in this case for the understanding of the tasks to be
performed and for the generation of data to understand the features
that are relevant to the task. One active research direction based
on this work is the automatic drilling of the mouse skull using AI-based
image recognition \cite{zhao2023autonomousdrilling}, in which the
completion level of the task is estimated through changes in the image.
The mouse skull is more challenging than the eggshell given that it
must be kept humid, making it difficult to discern where the drill
has passed. In addition, the mouse cranium is considerably flexible
owing to its thickness. Nonetheless, in the case of the mouse, anatomical
structures in the cranium, such as the cranial sutures, can be important
features for image recognition and AI-based control. In this case,
an in-depth study of experiments in mock setups, such as bimanual
handling in the peg-transfer task, the understanding of flexible material
handling in the gauze-cutting, and the drilling experiments on eggshells
will provide insights of motion that can be more readily achieved
by robotic systems without replicating human motion

We expect this platform to be the first step towards collaborative
AI robots that will autonomously perform scientific experiments that
otherwise could not be done by human scientists alone.

\bibliographystyle{IEEEtran}
\bibliography{bib/ram,bib/additional_references,bib/TeleRoboticSystems,bib/murilomarinho}

\end{document}